\documentclass[10pt,twocolumn,letterpaper]{article}

\usepackage{iccv}
\usepackage{times}
\usepackage{epsfig}
\usepackage{graphicx}
\usepackage{amsmath}
\usepackage{amssymb}

\usepackage[font=footnotesize,labelfont=bf]{caption}

\usepackage[belowskip=-5pt,aboveskip=3pt]{caption} 

\usepackage{multirow}
\usepackage{tabularx}
\usepackage{dsfont}
\usepackage{url}
\usepackage{color}
\usepackage{subcaption}
\usepackage{amsthm}
\usepackage[export]{adjustbox}
\usepackage{comment}

\usepackage[utf8]{inputenc} 
\usepackage[T1]{fontenc}    
\usepackage{url}            
\usepackage{booktabs}       
\usepackage{amsfonts}       
\usepackage{nicefrac}       
\usepackage{microtype}      
\usepackage{enumitem}
\usepackage{makecell}

\usepackage{multicol}

\def\argmin{\operatornamewithlimits{arg\,min}}

\definecolor{dg}{rgb}{0,0.694,0.298}
\definecolor{purple}{rgb}{0.4,0.176,0.569}
\definecolor{Remblue}{rgb}{0.368,0.745,0.941}

\usepackage{pifont}
\usepackage{xspace}
\makeatletter
\DeclareRobustCommand\onedot{\futurelet\@let@token\@onedot}
\def\@onedot{\ifx\@let@token.\else.\null\fi\xspace}
\def\eg{\emph{e.g}\onedot} 
\def\ie{\emph{i.e}\onedot} 
 
\def\etc{\emph{etc}\onedot} 
 
\def\etal{\emph{et al}\onedot}
\makeatother

\definecolor{royalblue}{RGB}{65,105,225} 
\usepackage[pagebackref=true,breaklinks=true,colorlinks,citecolor=royalblue,bookmarks=false]{hyperref}

\usepackage{colortbl}
\definecolor{tabgray}{rgb}{0.85,0.85,0.85}
\definecolor{top1}{rgb}{1.0, 0.6, 0.6}
\definecolor{top2}{rgb}{0.98, 0.91, 0.71}
\definecolor{top3}{rgb}{0.91, 1.0, 1.0}
\newcommand{\first}[1]{\textbf{\textcolor{red}{#1}}}
\newcommand{\second}[1]{\textbf{\textcolor{green}{#1}}}

\def\iccvPaperID{1406} 

\ificcvfinal\fi
\iccvfinalcopy

\begin{document}

\title{Learning to Adversarially Blur Visual Object Tracking}

\author{Qing Guo\textsuperscript{1,5}$^{*}$,
\,Ziyi Cheng\textsuperscript{2}\thanks{Qing Guo and Ziyi Cheng are co-first authors and contribute equally.},
\,\,Felix Juefei-Xu\textsuperscript{3},
Lei Ma\textsuperscript{4}$^\dagger$,
Xiaofei Xie\textsuperscript{5}\thanks{Lei Ma and Xiaofei Xie are corresponding authors (\href{mailto:malei@ait.kyushu-u.ac.jp}{ma.lei@acm.org},
    \href{xfxie@ntu.edu.sg}{xfxie@ntu.edu.sg}).},
\,Yang Liu\textsuperscript{5,6},
Jianjun Zhao\textsuperscript{2}\\~\\
\textsuperscript{1}\,College of Intelligence and Computing, Tianjin University, China,\\
\textsuperscript{2}\,Kyushu University, Japan,~~
\textsuperscript{3}\,Alibaba Group, USA,~~
\textsuperscript{4}\,University of Alberta, Canada,\\
\textsuperscript{5}\,Nanyang Technological University, Singapore,~~
\textsuperscript{6}\,Zhejiang Sci-Tech University, China
}
\maketitle

\begin{abstract}
Motion blur caused by the moving of the object or camera during the exposure can be a key challenge for visual object tracking, affecting tracking accuracy significantly.
In this work, we explore the robustness of visual object trackers against motion blur from a new angle, i.e., adversarial blur attack (ABA).
Our main objective is to online transfer input frames to their natural motion-blurred counterparts while misleading the state-of-the-art trackers during the tracking process.
To this end, we first design the motion blur synthesizing method for visual tracking based on the generation principle of motion blur, considering the motion information and the light accumulation process.
With this synthetic method, we propose \textit{optimization-based ABA (OP-ABA)} by iteratively optimizing an adversarial objective function against the tracking w.r.t. the motion and light accumulation parameters.
The OP-ABA is able to produce natural adversarial examples but the iteration can cause heavy time cost, making it unsuitable for attacking real-time trackers.
To alleviate this issue, we further propose \textit{one-step ABA (OS-ABA)} where we design and train a \textit{joint adversarial motion and accumulation predictive network (JAMANet)} with the guidance of OP-ABA, which is able to efficiently estimate the adversarial motion and accumulation parameters in a one-step way.
The experiments on four popular datasets (\eg, OTB100, VOT2018, UAV123, and LaSOT) demonstrate that our methods are able to cause significant accuracy drops on four state-of-the-art trackers with high transferability.
Please find the source code at \href{https://github.com/tsingqguo/ABA}{https://github.com/tsingqguo/ABA}.
\end{abstract}

\section{Introduction}\label{sec:intro}

\begin{figure}
\centering
\includegraphics[width=\linewidth]{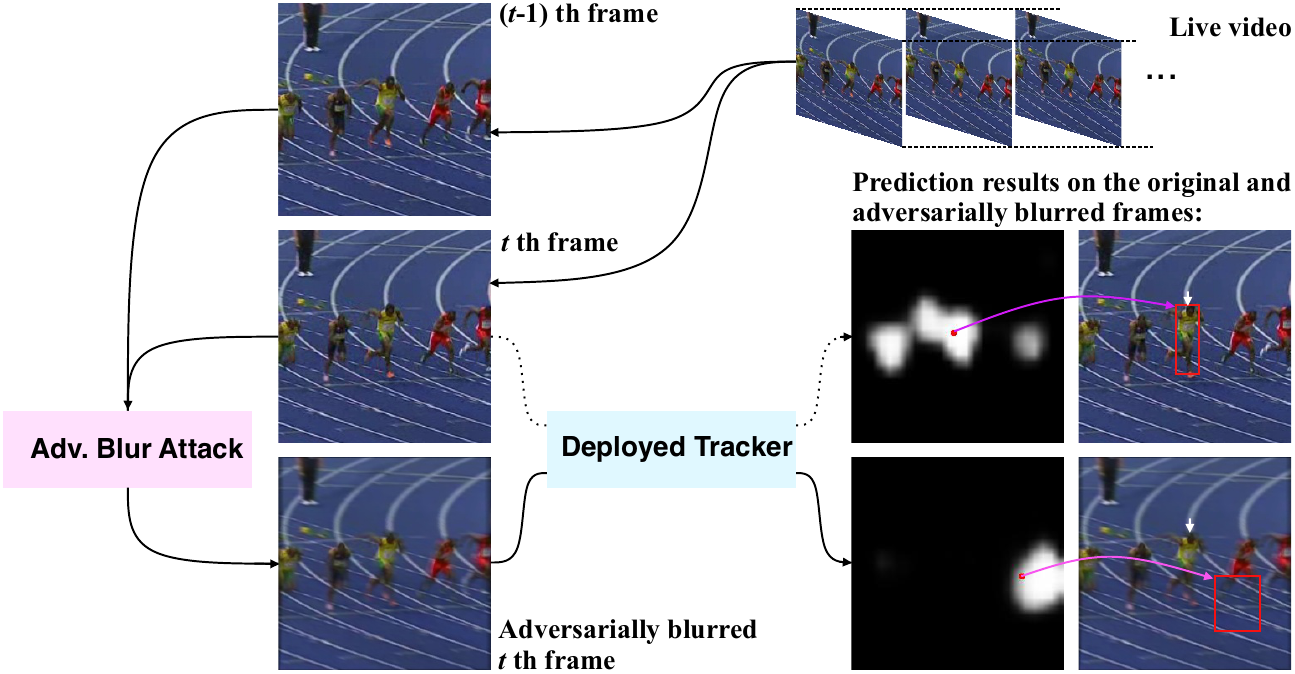}
   \caption{An example of our adversarial blur attack against a deployed tracker, \eg, SiamRPN++ \cite{Li2019CVPR}. Two adjacent frames are fed to our attack and it generates an adversarially blurred frame that misleads the tracker to output an inaccurate response map.}
\label{fig:idea}\vspace{-15pt}
\end{figure}
%
Visual object tracking (VOT) has played an integral part in multifarious computer vision applications nowadays ranging from augmented reality \cite{azuma2001recent,wei2019instant} to video surveillance \cite{hu2004survey}, from human-computer interaction \cite{munro2020multi,li2020tea} to traffic control \cite{wojek2012monocular}, \etc. Since the infusion of deep learning, VOT has become more powerful in terms of both algorithmic performance and efficiency \cite{guo2020selective}, leading to the more pervasive deployment of VOT-enabled on-device applications. However, the VOT can still exhibit robustness brittleness when faced with less ideal video feed. Among many known degrading factors such as illumination variations, noise variations, \etc, motion blur is perhaps one of the most important adverse factors for visual object tracking, which is caused by the moving of the object or camera during the exposure, and can severely jeopardize tracking accuracy \cite{guo2021exloring}.
Most of the existing benchmarks \cite{kristan2018sixth,Wu15,Mueller2016ECCV} only indicate whether a video or a frame contains motion blur or not and this piece of information is still insufficient to analyze the influence from motion blur by means of controlling all the variables, \eg, eliminating other possible interference from other degradation modes, which may lead to incomplete conclusions regarding the effects of motion blur in these benchmarks.

Moreover, the currently limited datasets, albeit being large-scale, cannot well cover the diversity of motion blur in the real world because motion blur is caused by camera and object moving in the scene which is both dynamic and unknown. Existing motion blur generation methods cannot thoroughly reveal the malicious or unintentional threat to visual object tracking, \ie, they can only produce natural motion blur which falls short of exposing the adversarial brittleness of the visual object tracker.
As a result, it is necessary to explore a novel motion blur synthetic method for analyzing the robustness of the visual object trackers, which should not only generate natural motion-blurred frames but also embed maliciously adversarial or unintentional threats.

In this work, we investigate the robustness of visual trackers against motion blur from a new angle, that is, adversarial blur attack (ABA). Our main objective is to online transfer input frames to their natural motion-blurred counterparts while misleading the state-of-the-art trackers during the tracking process. We show an intuitive example in Fig.~\ref{fig:idea}. To this end, we first design the motion blur synthesizing method for visual tracking based on the generation principle of motion blur, considering the motion information and the light accumulation process. With this synthetic method, we further propose \textit{optimization-based ABA (OP-ABA)} by iteratively optimizing an adversarial objective function against the tracking w.r.t. the motion and light accumulation parameters.

The OP-ABA is able to produce natural adversarial examples but the iteration can lead to a heavy time-consuming process that is not suitable for attacking the real-time tracker. To alleviate this issue, we further propose \textit{one-step ABA (OS-ABA)} where we design and train a \textit{joint adversarial motion and accumulation predictive network (JAMANet)} with the guidance of OP-ABA, which is able to efficiently estimate the adversarial motion and accumulation parameters in a one-step way. The experiments on four popular datasets (\eg, OTB100, VOT2018, UAV123, and LaSOT) demonstrate that our methods are able to cause significant accuracy drops on four state-of-the-art trackers while keeping the high transferability. To the best of our knowledge, this is the very first attempt to study the adversarial robustness of VOT and the findings will facilitate future-generation visual object trackers to perform more robustly in the wild.


\section{Related Work}\label{sec:related}

{\bf Visual object tracking (VOT).}
VOT is an important task in computer vision. Recently, a great number of trackers, which extract features with convolutional neural networks (CNNs), are proposed and achieve amazing performance.

Among these works, Siamese network-based methods \cite{bertinetto2016fully,Fan2019CVPR,li2018high,Guo17_ICCV,Zhu2018ECCV,Wang2018CVPR,Zhang2019CVPR,voigtlaender2020siam} offline train Siamese networks and conduct online matching between search regions and the object template, which are significantly faster with high tracking performance. In particular, SiamRPN \cite{li2018high, Li2019CVPR} embed the regional proposal network \cite{ren2015faster} in the naive Siamese tracker \cite{bertinetto2016fully}, allowing high-efficiency estimation of the object's aspect ratio variation and achieving state-of-the-art tracking accuracy.
After that, some works use historical frames to online update tracking models.
For example, DiMP \cite{Bhat2019ICCV} collects past frames' features and online predict convolution kernels that can estimate object's position. Furthermore, PrDiMP \cite{Danelljan2020CVPR} improves the loss function with KL divergence and information entropy from the perspective of probability distribution. KYS \cite{bhat2020know} considers the correlation between previous frames and the current frame.
These trackers run beyond real time and get top accuracy on several benchmarks.
Although great progress has been achieved, there are few works studying their robustness to motion blur. In this work, we identify a new way to achieve this goal by actively synthesizing adversarially motion blur to fool the state-of-the-art trackers.

{\bf Motion blur synthesis.}
In VOT task, motion blur is a very common scene due to the high-speed movement of the target. It is usually used to evaluate the quality of the trackers \cite{Wu15,Fan2019LaSOT,guo2021exloring}. In recent years, motion blur synthesis has been extensively studied in the rendering community \cite{navarro2011motion,guo2021exloring}. However, these methods usually require a complete understanding of the speed and depth of the scene as input. In order to get more realistic and high-quality images with motion blur, Brooks \etal \cite{Brooks2019CVPR} identify a simple solution that warps two instant images by optical flow \cite{sun2018pwc,ilg2017flownet} and fuses these intermediate frames with specific weights, to synthesize a blur picture.
This method is to synthesize realistic motion blur for the deblurring task while our work is used for adversarially blurring the frames for tracking.
Another related work, \ie, ABBA \cite{Guo2020NeurIPS}, takes a single image as its input and generates visually natural motion-blurred adversarial example to fool the deep neural network-based classification. Specifically, ABBA simulates the motion by adversarially shifting the object and background, respectively, neglecting the real motion in the scene. Different from ABBA, our approach focuses on visual object tracking with real object movement indicated by two adjacent frames.
Recently, some techniques \cite{bhat2020know,Danelljan2020CVPR,voigtlaender2020siam} have been proposed to counter the interference of the environment. To this end, our method is proposed to better evaluate the robustness of these VOTs.
\begin{figure*}
\centering
\includegraphics[width=0.95\linewidth]{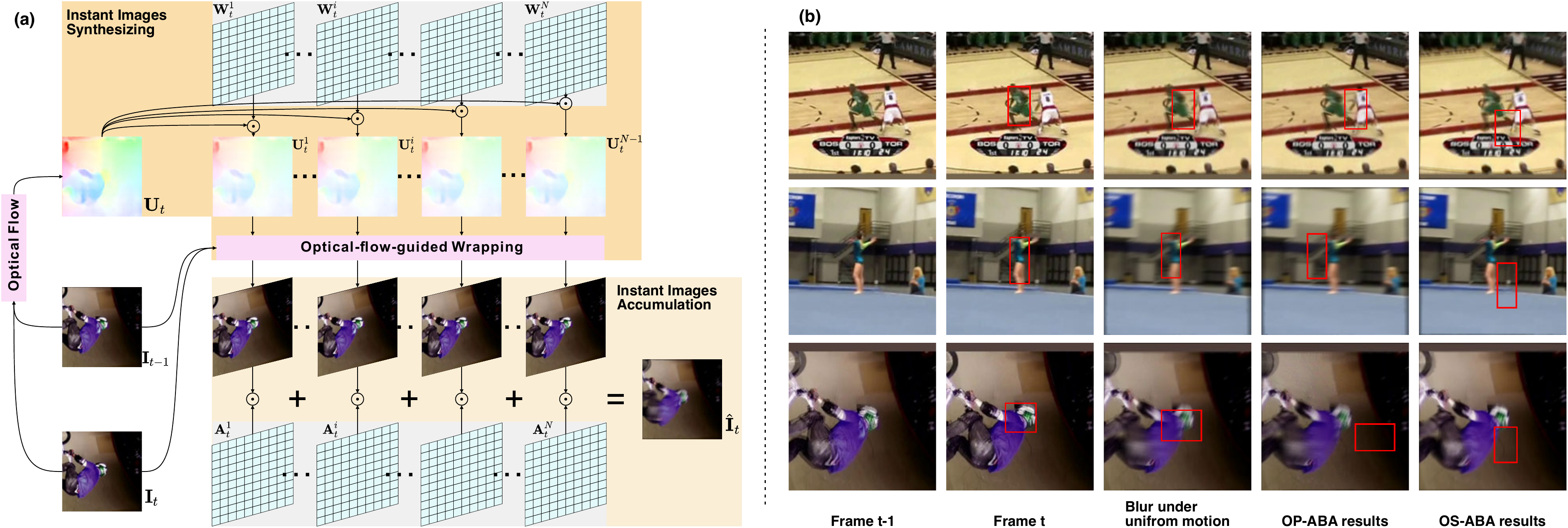}
   \caption{(a) shows the motion blur synthesizing process with two frames, \ie, $\mathbf{I}_t$ and $\mathbf{I}_{t-1}$, and two sets of variables, \ie, $\{\mathcal{A}_t^i\}$ and $\{\mathcal{W}_t^i\}$, should be determined for attacking. (b) shows three cases of the normal blur under uniform motion, the OP-ABA blurring results, and OS-ABA blurring results.}
\label{fig:blursys}
\end{figure*}

{\bf Adversarial attack.}
Extensive works have proved that state-of-the-art deep neural networks are still vulnerable to adversarial attacks by adding visually imperceptible noises or natural degradation to original images \cite{goodfellow2014explaining,tian2020bias,cheng2021pasadena, gao2020making,Guo2020NeurIPS}.
%
%
FGSM \cite{goodfellow2014explaining} perturbs normal examples along the gradient direction via the fast gradient sign method. MI-FGSM \cite{dong2018boosting} integrates momentum term into the iterative process that can help stabilize the update directions. C\&W \cite{carlini2017towards} introduces three new attacks for different norms ($L_0, L_2, L_\infty$) through iterative optimization. However, the above methods are unable to meet the real-time requirements due to the limited speed \cite{Guo2020ECCV}. To realize the efficient attacking, \cite{yan2020cooling,xiao2018generating} propose one-step attacks by offline training on the targeted model.
However, these methods are designed for the classification task and could not attack trackers directly.

More recently, some works have been proposed to attack visual object tracking. PAT \cite{wiyatno2019physical} generates physical adversarial textures via a white-box attack. SPARK \cite{Guo2020ECCV} studies how to adapt existing adversarial attacks on tracking. Chen \etal \cite{chen2020one} propose to add adversarial perturbations on the template at the initial frame. CSA \cite{yan2020cooling} raises a one-step method and makes objects invisible to trackers by forcing the predicted bounding box to shrink.
Different from above works, we employ motion blur to perform adversarial attack. Our work is designed to address three challenges: how to synthesize natural motion blur that meets the motion of object and background in the video; how to make the blurred frame fool the state-of-the-art trackers easily; how to perform the adversarial blur attack efficiently.
To the best of our knowledge, this is the very first attempt in the community of adversarial attack.

%
\section{Adversarial Blur Attack against Tracking}\label{sec:method}
%
%
In this section, we first study how to synthesize natural motion blur under the visual tracking task in Sec.~\ref{subsec:aba_tracking} and summarize the variables that should be solved to perform attacks. Then, we propose the optimization-based ABA (OP-ABA) in Sec.~\ref{subsec:op-aba} with a novel objective function to guide the motion blur generation via the iterative optimization process.
To allow high-efficiency attack for real-time trackers, we further propose one-step ABA (OS-ABA) in Sec.~\ref{subsec:os-aba} by training a newly designed \textit{joint motion and kernel predictive network} under the supervision of the objective function of OP-ABA.
Finally, we summarize the attacking details with OP-ABA and OS-ABA in Sec.~\ref{subsec:attack-alg}.

\subsection{Motion Blur Synthesizing for Visual Tracking}
\label{subsec:aba_tracking}
In a typical tracking process, given the $t$-th frame $\mathbf{I}_t$ of a live video and an object template specified at the first frame, a tracker uses a pre-trained deep model $\phi_{\theta_t}(\mathbf{I}_t)$ to predict the location and size of the object (\ie, the bounding box tightly wrapping the object) in this frame where $\theta_t$ denotes the template-related parameter and can be updated during the tracking process.
For the adversarial blur attack, we aim to generate a motion-blurred counterpart of $\mathbf{I}_t$, which is able to fool the tracker to estimate the incorrect bounding box of the object while having the natural motion-blur pattern.

To this end, we review the generation principle of realistic motion blur \cite{Nah2017CVPR,Noroozi2017,Brooks2019CVPR,Guo2020NeurIPS,guo2021exloring}: the camera sensor captures an image by receiving and accumulating light during the shutter procedure. The light at each time can be represented as an instant image, and there are a series of instant images for the shutter process. When the objects or background move, the light accumulation will cause blurry effects, which can be approximated by averaging the instant images.

Under the above principle, when we want to adversarially blur $\mathbf{I}_t$, we need to do two things: \textit{First}, synthesizing the instant images during the shutter process and letting them follow the motion of object and background in the video; \textit{Second}, accumulating all instant images to get the motion-blurred $\mathbf{I}_t$.
The main challenge is how to make the two steps adversarially tunable to fool the tracker easily while preserving the natural motion blur pattern.

For the first step, we propose to generate the instant images under the guidance of the optical flow $\mathbf{U}_t$ that describes the pixel-wise moving distance and direction between $\mathbf{I}_t$ and its neighbor $\mathbf{I}_{t-1}$.
Specifically, given two neighboring frames in a video, \eg, $\mathbf{I}_{t-1}$ and $\mathbf{I}_{t}$, we regard them as the start and end time stamps for camera shutter process, respectively. Assuming there are $N$ instant images, we denote them as $\{\mathbf{I}_{t}^{i}\}_{i=1}^{N}$ where $\mathbf{I}_{t}^{1}=\mathbf{I}_{t-1}$ and $\mathbf{I}_{t}^{N}=\mathbf{I}_{t}$.
Then, we calculate the optical flow $\mathbf{U}_{t}$ between $\mathbf{I}_{t-1}$ and $\mathbf{I}_{t}$ and split it into $N-1$ sub-motions, \ie, $\{\mathbf{U}_{t}^{i}\}_{i=1}^{N-1}$ where $\mathbf{U}_{t}^{i}$ represents the optical flow between $\mathbf{I}_{t}^{i}$ and $\mathbf{I}_{t}^{i+1}$. We define $\mathbf{U}_{t}^{i}$ as a scaled $\mathbf{U}_{t}$ with pixel-wise ratios (\ie, $\mathbf{W}_{t}^i$)
\begin{align} \label{eq:sub_opticalflow}
\mathbf{U}_{t}^{i}&=\mathbf{W}_{t}^i\odot\mathbf{U}_{t},
\end{align}
where $\mathbf{W}_{t}^i$ has the same size with $\mathbf{U}_{t}$ and $\odot$ denotes the pixel-wise multiplication.
All elements in $\mathbf{W}_{t}^i$ range from zero to one and we constraint the summation of $\{\mathbf{W}_{t}^i\}_{i=1}^{N}$ at the same position to be one, \ie,  $\forall \mathbf{p}, \sum_i^{N-1}\mathbf{W}_{t}^i[\mathbf{p}]=1$ where $\mathbf{W}_{t}^i[\mathbf{p}]$ denotes the $\mathbf{p}$-th element in $\mathbf{W}_{t}^i$.
Note that, the ratio matrices, \ie, $\{\mathbf{W}_{t}^i\}_{i=1}^{N}$, determine the motion pattern. For example, if we have $\forall\mathbf{p},\{\mathbf{W}_{t}^i[\mathbf{p}]=\frac{1}{N-1}\}_{i=1}^{N-1}$ and can calculate the sub-motions by $\{\mathbf{U}_{t}^{i}=\frac{1}{N-1}\mathbf{U}_{t}\}_{i=1}^{N-1}$, all pixels follow the uniform motion.

With Eq.~\eqref{eq:sub_opticalflow}, we get all sub-motions (\ie, $\{\mathbf{U}_t^i\}_{i=1}^{N-1}$) and produce the instant images by warping $\mathbf{I}_{t}$ w.r.t. different optical flows. For example, we synthesize $\mathbf{I}_t^i$ by
{\small
\begin{align} \label{eq:instantimages}
\mathbf{I}_t^i = \frac{1}{2}\text{warp}(\mathbf{I}_{t-1},\sum_{j=1}^{i-1}\mathbf{W}_{t}^j\odot\mathbf{U}_{t}^{j}) +
 \frac{1}{2}\text{warp}(\mathbf{I}_{t},\sum_{j=i}^{N-1}\mathbf{W}_{t}^j\odot\mathbf{U}_{t}^{j}),
\end{align}}
where $\sum_{j=1}^{i-1}\mathbf{W}_{t}^j\odot\mathbf{U}_{t}^{j}$ represents the optical flow between $\mathbf{I}_{t-1}$ and $\mathbf{I}_t^{i}$ while $\sum_{j=i}^{N-1}\mathbf{W}_{t}^j\odot\mathbf{U}_{t}^{j}$ denotes the optical flow between  $\mathbf{I}_{t}^i$ and $\mathbf{I}_t$. The function $\text{warp}(\cdot)$ is to warp the $\mathbf{I}_{t-1}$ or  $\mathbf{I}_{t}$ according to the corresponding optical flow, and uses the implementation in \cite{Guo2020NeurIPS} for spatial transformer network.

For the second step, after getting $\{\mathbf{I}_t^i\}_{i=1}^{N}$, we can synthesize the motion-blurred $\mathbf{I}_t$ by summing up the $N$ instant images with pixel-wise accumulation weights $\{\mathbf{A}_i\}_{i=1}^{N}$
\begin{align} \label{eq:syn_motionblur}
\hat{\mathbf{I}}_t = \sum_{i=1}^{N}\mathbf{A}_{t}^i\odot\mathbf{I}_t^i.
\end{align}
where $\mathbf{A}_{t}^i$ has the same size with $\mathbf{I}_t^i$ and all elements range from zero to one. For simulating realistic motion blur, all elements of $\mathbf{A}_{t}^i$ are usually fixed as $\frac{1}{N}$, which denotes the accumulation of all instant images.

Overall, we represent the whole blurring process via Eqs.~\eqref{eq:syn_motionblur} and \eqref{eq:instantimages} as
$\hat{\mathbf{I}}_t=\text{Blur}(\mathbf{I}_t,\mathbf{I}_{t-1},\mathcal{W}_t,\mathcal{A}_t).$
To perform adversarial blur attack for the frame $\mathbf{I}_t$, we need to solve two sets of variables, \ie, $\mathcal{W}_t=\{\mathbf{W}_{t}^i\}_{i=1}^{N-1}$ determining the motion pattern and $\mathcal{A}_t=\{\mathbf{A}_{t}^i\}_{i=1}^{N}$ deciding the accumulation strategy.
In Sec.~\ref{subsec:op-aba}, we follow the existing adversarial attack pipeline and propose the optimization-based ABA by defining and optimizing a tracking-related objective function to get $\mathcal{W}_t$ and $\mathcal{A}_t$.
In Sec.~\ref{subsec:os-aba}, we design a network to predict $\mathcal{W}_t$ and $\mathcal{A}_t$ in a one-step way.

\subsection{Optimization-based Adversarial Blur Attack}
\label{subsec:op-aba}
In this section, we propose to solve $\mathcal{W}_t$ and $\mathcal{A}_t$ by optimizing the tracking-related objective function.
Specifically, given the original frame $\mathbf{I}_t$, a tracker can estimate a response or classification map by $
\mathbf{Y}_t = \phi_{\theta_t}(\mathbf{I}_t)$ whose maximum indicates the object's position in the $\mathbf{I}_t$.
Our attack aims to generate a blurred $\mathbf{I}_t$ (\ie, $\hat{\mathbf{I}}_t=\text{Blur}(\mathbf{I}_t,\mathbf{I}_{t-1},\mathcal{W}_t,\mathcal{A}_t)$) to let the predicted object position indicated by $\hat{\mathbf{Y}}_t = \phi_{\theta_t}(\hat{\mathbf{I}}_t)$ be far away from the original one indicated by $\mathbf{Y}_t$.

To this end, we optimize $\mathcal{W}_t$ and $\mathcal{A}_t$ by minimizing
\begin{align}\label{eq:adv_obj}
\argmin_{\mathcal{W}_t, \mathcal{A}_t}& J(\phi_{\theta_t}(\text{Blur}(\mathbf{I}_t,\mathbf{I}_{t-1},\mathcal{W}_t,\mathcal{A}_t)),\mathbf{Y}^*_t) \nonumber \\
\mathrm{~~subject~to~~}& \forall\mathbf{p}, \forall i, \sum_i^{N-1}\mathbf{W}_{t}^i[\mathbf{p}]=1, \sum_i^{N}\mathbf{A}_{t}^i[\mathbf{p}]=1,
\end{align}
where the two constraints on $\mathbf{W}_t^i$ and $\mathbf{A}_t^i$ make sure the synthetic motion blur does not have obvious distortions.
The function $J(\cdot)$ is a distance function and is set as $L_2$.
The regression target $\mathbf{Y}^{*}_t$ denotes the desired response map and is obtained under the guidance of the original $\mathbf{Y}_t$. Specifically, with the original response map $\mathbf{Y}_t$, we know the object's position and split $\mathbf{Y}_t$ into two regions the object region and background region according to the object size. Then, we can find the position (\eg, $\mathbf{q}$) having the highest response score at the background region of $\mathbf{Y}_t$ and then we set $\mathbf{Y}^{*}[\mathbf{q}]=1$ and other elements of $\mathbf{Y}^{*}$ to be zero.
Note that, the above setup is suitable for regression-based trackers, \eg, DiMP and KYS, and can be further adapted to attack classification-based trackers, \eg, SiamRPN++, by setting $J(\cdot)$ as the cross-entropy loss function and $\mathbf{Y}^*[\mathbf{q}]=1$ with its other elements to be $-1$.

Following the common adversarial attacks \cite{Goodfellow_2014_arxiv,Dong2018BoostingAA,Guo2020ECCV,Guo2020NeurIPS}, we can solve Eq.~\eqref{eq:adv_obj} via the signed gradient descent and update the $\mathcal{W}_t$ and $\mathcal{A}_t$ iteratively with specified step size and iterative number. We show the synthesized motion blur of OP-ABA in Fig.~\ref{fig:blursys}. Clearly, OP-ABA is able to synthesize natural motion-blurred frames that have a similar appearance to the normal motion blur.
%

\begin{figure*}[t]
\centering
\includegraphics[width=0.7\linewidth]{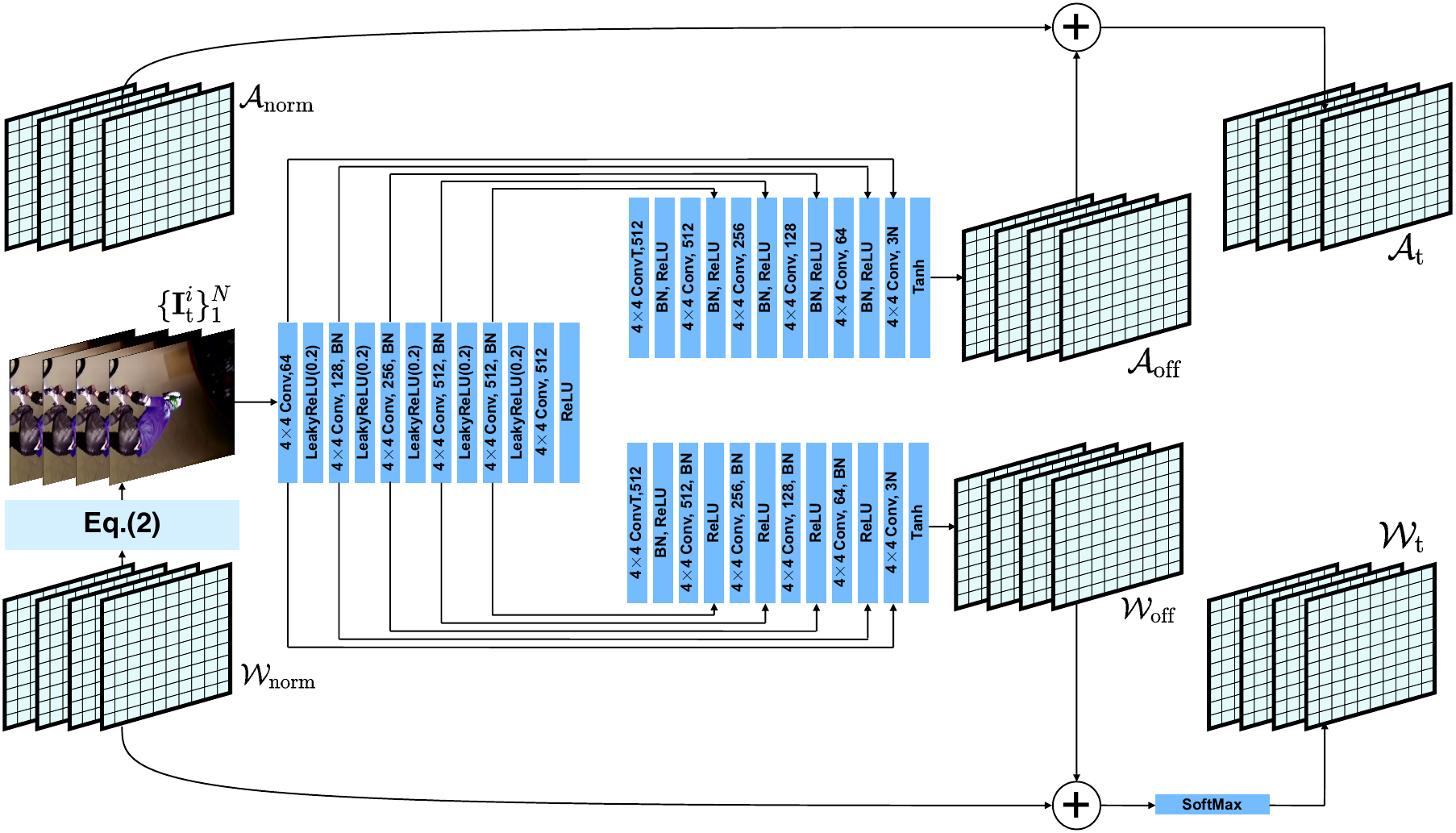}
   \caption{Architecture of JAMANet.}
\label{fig:jamanet}
\end{figure*}

\subsection{One-Step Adversarial Blur Attack}
\label{subsec:os-aba}
To allow efficient adversarial blur attack, we propose to predict the motion and accumulation weights (\ie, $\mathcal{W}_t$ and $\mathcal{A}_t$) with a newly designed network denoted as \textit{joint adversarial motion and accumulation predictive network (JAMANet)} in a one-step way, which is pre-trained through the objective function Eq.~\eqref{eq:adv_obj} and a naturalness-aware loss function.
Specifically, we use JAMANet to process the neighboring frames (\ie, $\mathbf{I}_t$ and $\mathbf{I}_{t-1}$) and predict the $\mathcal{W}_t$ and $\mathcal{A}_t$, respectively.
Meanwhile, we also employ a pre-trained network to estimate the optical flow $\mathbf{U}_t$ between $\mathbf{I}_t$ and $\mathbf{I}_{t-1}$. Here, we use the PWCNet \cite{sun2018pwc} since it achieves good results on diverse scenes.
Then, with Eq.~\eqref{eq:instantimages}-\eqref{eq:syn_motionblur}, we can obtain the motion-blurred frame $\hat{\mathbf{I}}_t$.
After that, we feed $\hat{\mathbf{I}}_t$ into the loss functions and calculate gradients of parameters of JAMANet to perform optimization.
We show the framework in Fig.~\ref{fig:jamanet}.

{\bf Architecture of JAMANet.}
We first build two parameter sets with constant values, which are denoted as $\mathcal{A}_\text{norm}=\{\mathbf{A}_\text{norm}^i\}$ and $\mathcal{W}_\text{norm}=\{\mathbf{W}_\text{norm}^i\}$.
All elements in $\mathcal{A}_\text{norm}$ and $\mathcal{W}_\text{norm}$ are fixed as $\frac{1}{N}$ and $\frac{1}{N-1}$, respectively.
We then use $\mathcal{W}_\text{norm}$, $\mathbf{I}_{t-1}$, and $\mathbf{I}_{t}$ to generate $N$ instant images through Eq.~\eqref{eq:instantimages}.
JAMANet is built based on the U-Net architecture \cite{ronneberger2015u} but contains two decoder branches, which is fed with the $N$ instant images $\{\mathbf{I}_t^i\}_{i=1}^{N}$ and outputs the offsets w.r.t. $\mathcal{W}_\text{norm}$ and $\mathcal{A}_\text{norm}$. We name them as $\mathcal{W}_\text{off}$ and $\mathcal{A}_\text{off}$.
The input $\{\mathbf{I}_t^i\}_{i=1}^{N}$ is size of $(N,3,H,W)$.
We resize it to $(1,3N,H,W)$ and normalize the values to the range of -1 to 1.
The architecture is a fully convolutional encoder/decoder model with skip connections. In encoder stage, we use six convolutions with the kernel size 4x4 and the LeakyReLU \cite{maas2013rectifier} activation function. Unlike the standard U-Net, JAMANet has two decoders. Specifically, one branch is set to estimate the $\mathcal{A}_\text{off}$, containing six transposed convolutions \cite{zeiler2011adaptive} with the latest activation function as $\mathrm{Tanh}$.
We can calculate the final $\mathcal{A}_t$ through $\mathcal{A}_t=\mathcal{A}_\text{norm}+\mathcal{A}_\text{off}$.
Another branch is to predict $\mathcal{W}_\text{off}$ and get $\mathcal{W}_t = \mathcal{W}_\text{norm}+\mathcal{W}_\text{off}$. This architecture is the same with the previous one but following a $\mathrm{Softmax}$ for catering to constraints of Eq.~\eqref{eq:adv_obj}\footnote{\scriptsize To let $\mathcal{A}_t$ also meet the constraints, for any pixel $\mathbf{p}$, we first select the element $j=\argmin_{i,i\in [1,N]}{\mathbf{A}_\text{off}^{i}[\mathbf{p}]}$ and then set $\mathbf{A}_t^{j}[\mathbf{p}]=1-\sum_{i,i\neq j}^{N}\mathbf{A}_t^i[\mathbf{p}]$. }.

{\bf Loss functions.} We train the JAMANet with two loss functions:
\begin{align} \label{eq:lossblur1}
\mathcal{L} = \mathcal{L}_\text{adv}+\lambda\mathcal{L}_\text{natural},
\end{align}
where the first loss function, \ie, $\mathcal{L}_\text{adv}$, is set to the objective function in Eq.~\eqref{eq:adv_obj} to make sure the background content instead of the object be highlighted.
%
%
Note that, this loss function means to enhance the capability of adversarial attack, that is, misleading the raw trackers.
It, however, neglects the naturalness of adversarial blur.
To this end, we set the loss function $\mathcal{L}_\text{natural}$ as
\begin{align} \label{eq:lossblur3}
\mathcal{L}_\text{natural} = \sum_i^{N}\|\mathbf{A}_t^i-\mathbf{A}_\text{norm}^i\|_2.
\end{align}
This loss function encourages the estimated accumulation parameters to be similar to the normal ones, leading to natural motion blur.

{\bf Training details.} We use GOT-10K \cite{huang2019got} as our training
dataset, which includes 10,000+ sequences and 500+ object classes. For each video in GOT-10K \cite{huang2019got}, we set the first frame as template and take two adjacent frames as an image pair, \ie, $(\mathbf{I}_{t-1},\mathbf{I}_t)$. We select eight image pairs from each video. The template and two adjacent frames make up a training sample.
Here, we implement the OS-ABA for attacking two trackers, \ie, SiamRPN++ \cite{Li2019CVPR} with ResNet50 and MobileNetv2, respectively. In the experiment, we show that OS-ABA has strong transferability against other state-of-the-art trackers.
During the training iteration, we first calculate the template's embedding to construct tracking model $\phi_{\theta_t}$ and the original response map $\mathbf{Y}_t$ (\ie, the positive activation map of SiamRPN++). Then, we get $\mathbf{Y}^*_t$ and initialize the blurred frame via $\text{Blur}(\mathbf{I}_t, \hat{\mathbf{I}}_{t-1})$.
We can calculate the loss via Eq.~\eqref{eq:lossblur1} and obtain the gradients of the JAMANet via backpropagation for parameter updating.
We train the JAMANet for 10 epochs, requiring a total of about 9 hours on a single Nvidia RTX 2080Ti GPU. We use the Adam \cite{Kingma2015adam} with the learning rate of 0.0002 to optimize network parameters, and the loss weight $\lambda$ equals 0.001.

\begin{table*}[tbp]
\centering
\scriptsize
\caption{Attacking results of OP-ABA and OS-ABA against SiamRPN++ with ResNet50 and MobileNetv2 on OTB100 and VOT2018. The best results are highlighted by \first{red} color.} \label{tab:valid_1}
\begin{tabular}{l|c|c|c|c|c|c|c}
\toprule
    \multirow{2}{*}{SiamRPN++} & \multirow{2}{*}{Attacks}  &\multicolumn{4}{c|}{OTB100} & \multicolumn{2}{c}{VOT2018} \\
       &  &  Org.~Prec. & Prec.~Drop $\uparrow$ & Org.~Succ. & Succ.~Drop $\uparrow$ & Org. EAO  & EAO~Drop $\uparrow$ \\
\midrule
    \multirow{2}{*}{ResNet50} & OP-ABA  & 87.8 & \first{41.7} & 66.5 & \first{31.2} & 0.415 & \first{0.375} \\
    & OS-ABA & 87.8 & 32.5 & 66.5 & 28.1 & 0.415 & 0.350 \\
\midrule
    \multirow{2}{*}{MobNetv2} & OP-ABA  & 86.4 & \first{49.6} & 65.8 & \first{37.6} & 0.410 & \first{0.384} \\
    & OS-ABA & 86.4 & 37.3 & 65.8 & 30.1 & 0.410 & 0.338 \\
\bottomrule
\end{tabular}
\end{table*}

\begin{table*}[tbp]
\centering
\scriptsize
\caption{Attacking results of OP-ABA and OS-ABA against SiamRPN++ with ResNet50 and MobileNetv2 on UAV123 and LaSOT. The best results are highlighted by \first{red} color.} \label{tab:valid_2}
\begin{tabular}{l|c|c|c|c|c|c|c|c|c}
\toprule
    \multirow{2}{*}{SiamRPN++} & \multirow{2}{*}{Attacks}  & \multicolumn{4}{c|}{UAV123} & \multicolumn{4}{c}{LaSOT}   \\
       &  & Org. Prec. & Prec.~Drop. $\uparrow$ & Org.~Succ. & Succ.~Drop $\uparrow$ &  Org.~Prec. & Prec.~Drop $\uparrow$ & Org.~Succ. & Succ.~Drop $\uparrow$ \\
\midrule
    \multirow{2}{*}{ResNet50} & OP-ABA  & 80.4 & \first{30.4} & 61.1 & \first{23.1} &  49.0 & \first{28.7} & 49.7 & 25.2 \\
    & OS-ABA & 80.4 & 29.6 & 61.1 & 19.9 & 49.0 & 26.8 & 49.7 & \first{26.4} \\
\midrule
    \multirow{2}{*}{MobNetv2} & OP-ABA & 80.2 & \first{34.7} & 60.2 & \first{26.9} &  44.6 & \first{29.7} & 44.7 & \first{28.1}\\
    & OS-ABA & 80.2  & 31.9 & 60.2 & 24.0 & 44.6 & 22.5 &  44.7 & 18.7 \\
\bottomrule
\end{tabular}
\end{table*}

\subsection{Attacking Details}
\label{subsec:attack-alg}
Intuitively, given a targeted tracker, we can attack it by blurring each frame through OP-ABA and OS-ABA during the online tracking process, as shown in Fig.~\ref{fig:idea}.
The attack could be white-box, that is, the tracking model in Eq.~\eqref{eq:adv_obj} is the same as the targeted tracker, leading to high accuracy drop.
It also could be black-box also known as the transferability, that is, the tracking model Eq.~\eqref{eq:adv_obj} is different from the targeted one.
Note that, OP-ABA is based on iterative optimization and is time-consuming, thus we conduct OP-ABA every five frames while performing OS-ABA for all frames.
In practice, we perform the blurring on the search regions between two frames to accelerate the attacking speed.
Specifically, at the frame $t$, we crop a search region centered at the detected object as the $\mathbf{I}_t$. At the same time, we crop a region from the previous frame at the same position as the $\mathbf{I}_{t-1}$.
Then, we use the PWCNet \cite{sun2018pwc} to calculate optical flow.
We get the original response map with the targeted tracker and $\mathbf{I}_t$ if we employ the OP-ABA as the attack method.
After that, we can conduct the OP-ABA or OS-ABA to generate the adversarial blurred frame.
In terms of the OP-ABA, we set the iteration number to be 10 and the step sizes for updating $\mathcal{W}_t$ and $\mathcal{A}_t$ are set as 0.002 and 0.0002, respectively. The number of intermediate frames $N$ is fixed as 17 for both OP-ABA and OS-ABA.

\section{Experimental Results}\label{sec:experiments}
We design experiments to investigate three aspects: \textit{First}, we validate the effectiveness of our two methods against state-of-the-art trackers on four public tracking benchmarks in Sec.~\ref{subsec:validation}. \textit{Second}, we design ablation experiments to validate the influences of $\mathcal{A}_t$ and $\mathcal{W}_t$ in Sec.~\ref{subsec:ablation}. \textit{Third}, we compare our method with the state-of-the-art tracking attacks on their transferability and frame quality in Sec.~\ref{subsec:comparion}.
\subsection{Setups}\label{sec:setup}
\textbf{Datasets.} We evaluated adversarial blur attack on four popular datasets, \ie, VOT2018 \cite{kristan2018sixth}, OTB100 \cite{Wu15}, UAV123 \cite{Mueller2016ECCV}, and LaSOT \cite{Fan2019LaSOT}. VOT2018 and OTB100 are widely used datasets containing 100 videos and 60 videos, respectively. LaSOT is a recent large-scale tracking benchmark, which contains 280 videos. UAV123 \cite{Mueller2016ECCV} focuses on tracking the object captured by unmanned aerial vehicle's camera, including 123 videos.

\textbf{Tracking models.}
We conduct attack against state-of-the-art trackers including SiamRPN++ \cite{Li2019CVPR} with ResNet50 \cite{he2016deep} and MobileNetv2 \cite{howard2017mobilenets}, DiMP \cite{Bhat2019ICCV} with ResNet50 and ResNet18, and KYS \cite{bhat2020know}.
Specifically, we validate the white-box attack with OP-ABA and OS-ABA against SiamRPN++ \cite{Li2019CVPR} with ResNet50 \cite{he2016deep} and MobileNetv2 in Sec.~\ref{subsec:validation} where the motion-blurred frames are guided by the targeted tracker's model itself.
We choose SiamRPN++ \cite{Li2019CVPR} since it is a classic tracker for Siamese network-based methods \cite{voigtlaender2020siam,li2018high,Dong2018ECCV,bertinetto2016fully,Guo17_ICCV} which achieves excellent tracking accuracy and real-time tracking speed.
We also conduct transferability experiments by using the motion blur crafted from SiamRPN++ with ResNet50 to attack other trackers.

\textbf{Metrics.}
In terms of the OTB100, UAV123, and LaSOT datasets, we follow their common setups and use one pass evaluation (OPE) that contains two metrics \textit{success rate} and \textit{precision}. The former one is based on the intersection over union (IoU) between the ground truth bounding box and predicted one for all frames while the latter is based on the center location error (CLE) between the ground truth and prediction. Please refer to \cite{Wu15} for details.
To evaluate the capability of attacking, we use the drop of success rate and precision for different attacks, which are denoted as Succ.~Drop and Prec.~Drop. The higher drops mean more effective attacking.
In terms of VOT2018, it restarts trackers when the object is lost. Expected average overlap (EAO) \cite{kristan2015visual} is the main criterion, evaluating both accuracy and robustness.
Similar to Succ.~Drop, we use the drop of EAO (\ie, EAO~Drop) for evaluating attacks.
When comparing with other additive noise-based attacks, we use the BRISQUE \cite{mittal2012noreference} as the image quality assessment. An attack is desired to produce adversarial examples that are not only natural but also able to fool trackers. BRISQUE is a common metric to evaluate the naturalness of images and a smaller BRISQUE means a more natural image.

\textbf{Baselines.}
There are several tracking attacks including cooling-shrinking attack (CSA) \cite{yan2020cooling}, SPARK \cite{Guo2020ECCV}, One-shot-based attack \cite{chen2020one}, and PAT \cite{wiyatno2019physical}. Among them, CSA and SPARK have released their code. We select CSA and SPARK as the baselines.

\begin{table}[t]
\centering
\scriptsize
\caption{Speed and time cost of three attacks and SiamRPN++ with the ResNet50 and MobileNetv2.} \label{tab:speed}
\begin{tabular}{c|c|c|c|c}
\toprule
  SiamRPN++ & Attackers & Org. FPS & \makecell{Attack time (ms) \\ per frame} $\downarrow$ & Attack FPS $\uparrow$  \\
 \midrule
 \multirow{2}{*}{ResNet50} & OP-ABA & 70.25 & 661.90 & 6.79  \\
  &  OS-ABA &  70.25  & \first{42.97} & \first{17.62} \\
 \midrule
 \multirow{2}{*}{MobNetv2} & OP-ABA & 107.62   & 508.30  &8.79  \\
  &  OS-ABA      & 107.62    & \first{40.88} & \first{19.96}  \\
\bottomrule
\end{tabular}
\end{table}

\subsection{Validation Results}
\label{subsec:validation}
%

\begin{figure*}[t]
\centering
\includegraphics[width=0.94\linewidth]{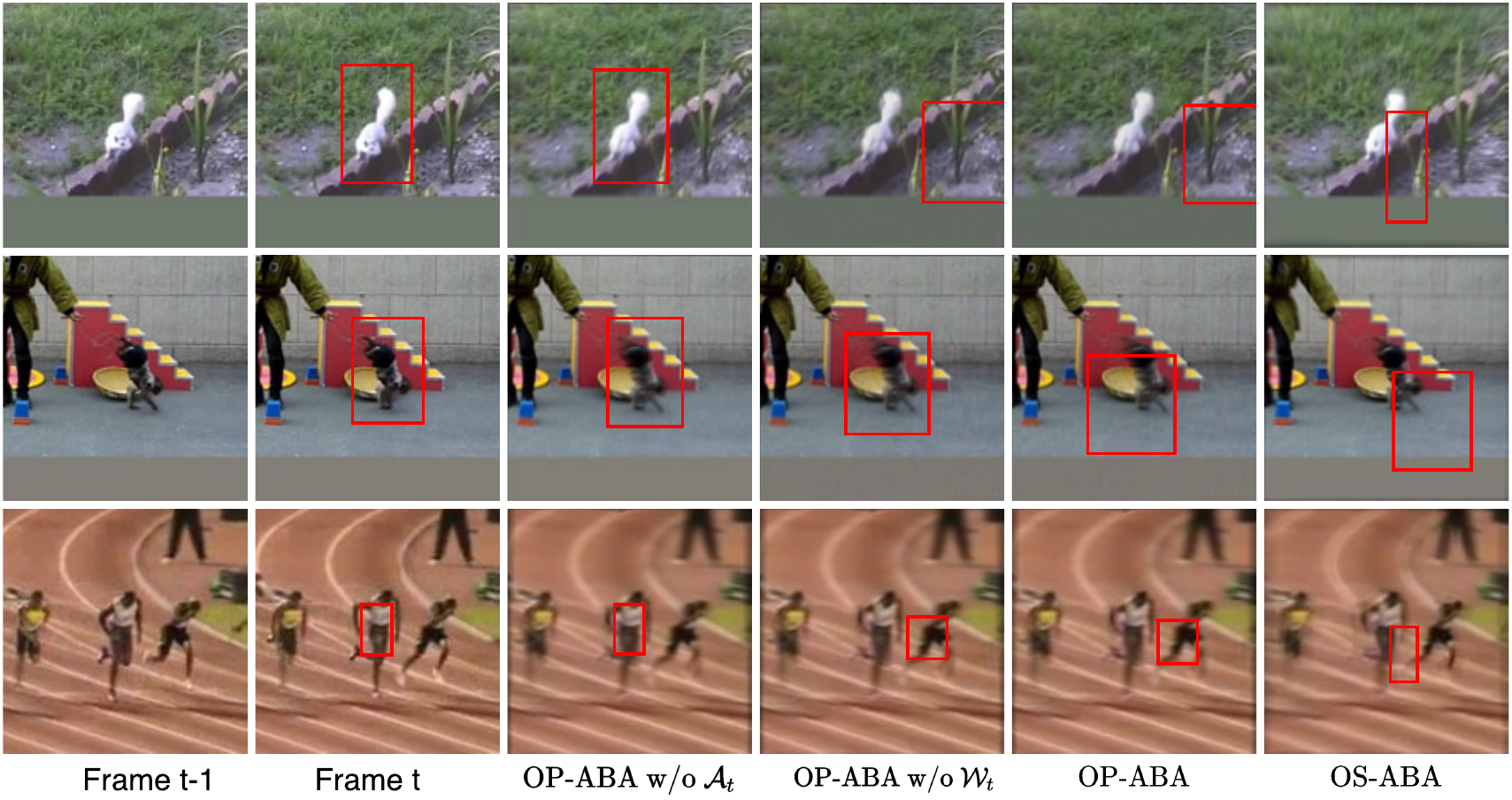}
   \caption{Three visualization results of OP-ABA w/o $\mathcal{A}_t$, OP-ABA w/o $\mathcal{W}_t$, OP-ABA, and OS-ABA against SiamRPN++~(ResNet50). The corresponding tracking results are showed with \first{red} bounding boxes.}
\label{fig:vis}\vspace{-5pt}
\end{figure*}

{\bf Attacking results.}
We attack two SiamRPN++ trackers that uses ResNet50 and MobileNetv2 as the backbone, respectively. The attacks results on the four public datasets are presented in Table~\ref{tab:valid_1} and \ref{tab:valid_2}, respectively.
We observe that: \ding{182} Both OP-ABA and OS-ABA reduce the success rate and precision of the two targeted trackers significantly on all benchmarks. Specifically, on the OTB100 dataset, OP-ABA makes the precision and success rate of SiamRPN++ with ResNet50 reduce 41.7 and 31.2, respectively, almost fifty percent of the original scores. These results demonstrate that the proposed attacks are able to fool the state-of-the-art trackers effectively. \ding{183} Compared with OS-ABA, OP-ABA achieves higher precision drop since it targeted attack to a certain position during each optimization while OS-ABA generates a general blurred image to make objects invisible for trackers. In general, all the results indicate the effectiveness of OP-ABA and OS-ABA in misleading the tracking models by adversarial blur attack. \ding{184} Comparing the performance drop of SiamRPN++ (ResNet50) with SiamRPN++ (MobileNetv2), we observe that the former usually has relatively smaller precision or success rate drop under the same attack, hinting that the lighter model is fooled more easily.  \ding{185} According to the visualization results shown in Fig.~\ref{fig:vis}, we see that both methods are able to generate visually nature blurred frames that mislead the SiamRPN++. In general, OP-ABA contains some artifacts but is able to mislead the tracker more effectively than OS-ABA. In contrast, OS-ABA always generates more realistic motion blur than OP-ABA in all three cases.


{\bf Speed analysis.}
We test the time cost of OP-ABA and OS-ABA on the OTB100 and report the FPS of the SiamRPN++ trackers before and after attacking.
As presented in Table \ref{tab:speed} , we observe that OP-ABA would slow down the tracking speed significantly. For example, OP-ABA reduces the speed of SiamRPN++ with ResNet-50 from 63 FPS to 6.79 PFS due to the online optimization. Thanks to the one-step optimization via JAMANet in Sec.~\ref{subsec:os-aba} , OS-ABA is almost ten times faster than OP-ABA according to the average attack time per frame. In consequence, OS-ABA achieved near real-time speed, \eg, 17.62 FPS and 20.00 FPS, in attacking SiamRPN++ (ResNet50) and SiamRPN++ (MobileNetv2). In terms of the FPS after attacking, OS-ABA also about 3 times faster than OP-ABA.

\subsection{Ablation Study}
\label{subsec:ablation}
%
\begin{table}[t]
\centering
\scriptsize
\caption{Effects of $\mathcal{W}_t$ and $\mathcal{A}_t$ to OP-ABA and OS-ABA by attacking SiamRPN++ (ResNet50) on OTB100. The best results are highlighted by \first{red} color.} \label{tab:ablation}
\begin{tabular}{l|c|c|c|c}
\toprule
  Attackers & Succ. Rate & Succ.~Drop $\uparrow$  &  Prec. & Prec.~Drop $\uparrow$ \\
\midrule
  Original                    & 66.5    & 0.0  & 87.8 & 0.0   \\
  Norm-Blur                   & 65.3    & 1.2  & 86.2 & 1.6 \\
\midrule
  OP-ABA w/o $\mathcal{A}_t$  & 51.5    & 15.0 & 67.6 & 20.2 \\
  OP-ABA w/o $\mathcal{W}_t$  & 40.9    & 25.6 & 53.4 & 34.4 \\
  OP-ABA                      & 35,3    & \first{31.2} & 46.1 & \first{41.7} \\
\midrule
  OS-ABA w/o $\mathcal{A}_t$  & 61.0    & 5.5  & 80.8 & 7.0 \\
  OS-ABA w/o $\mathcal{W}_t$  & 41.6    & 24.9 & 58.3 & 29.5 \\
  OS-ABA                      & 38.4    & \first{28.1} &55.3 & \first{32.5} \\
\bottomrule
\end{tabular}
\end{table}
%
In this section, we discuss the influence of $\mathcal{W}_t$ and $\mathcal{A}_t$ to OP-ABA and OS-ABA by constructing two variants of them to attack SiamRPN++~(ResNet50) tracker on OTB100 dataset. Specifically, for both attacks, we only tune $\mathcal{A}_t$ and fix $\mathcal{W}_t$ as $\mathcal{W}_\text{norm}$, thus we get two variants OP-ABA w/o $\mathcal{W}_t$ and OS-ABA w/o $\mathcal{W}_t$. Similarly, we replace $\mathcal{A}_t$ with $\mathcal{A}_\text{norm}$ and adversarially tune $\mathcal{W}_t$, thus we get OP-ABA w/o $\mathcal{A}_t$ and OS-ABA w/o $\mathcal{A}_t$, respectively.
Moreover, to demonstrate that it is the adversarial blur that reduces the performance, we build the `Norm-Blur' attack. It synthesizes the motion blur with $\mathcal{A}_\text{norm}$ and $\mathcal{W}_\text{norm}$, representing the normal blur that may appear in the real world.

We summarize the results in Table~\ref{tab:ablation} and Fig.~\ref{fig:vis} and have the following observations:
\ding{182} When we fix the $\mathcal{W}_t$ or $\mathcal{A}_t$ for OP-ABA and OS-ABA, the success rate and precision drops decrease significantly, demonstrating that tuning both motion pattern (\ie, $\mathcal{W}_t$ ) and accumulation strategy ($\mathcal{A}_t$) can benefit the adversarial blur attack. \ding{183} According to the variance of the performance drop, we see that tuning the accumulation strategy ($\mathcal{A}_t$) contributes more for effective attacks. For example, without tuning $\mathcal{A}_t$, the success rate drop reduces from 28.1 and 31.2 to 5.5 and 15.0 for OS-ABA and OP-ABA, respectively. \ding{184} SiamRPN++ are robust to the Norm-Blur with slight success rate and precision drops. In contrast, the adversarial blur causes a significant performance drop, demonstrating the adversarial blur does pose threat to visual object tracking. \ding{185} According to the visualization results in Fig.~\ref{fig:vis}, we have similar conclusions with the quantitative results in Table~\ref{tab:ablation}: OP-ABA w/o $\mathcal{A}_t$ can generate motion-blurred frames but have little influence on the prediction accuracy. Once we tune $\mathcal{A}_t$, the tracker can be fooled effectively but some artifacts are also introduced.

\subsection{Comparison with Other Attacks}
\label{subsec:comparion}
In this section, we study the transferability of proposed attacks by comparing them with baseline attacks, \ie, CSA \cite{yan2020cooling} and SPARK \cite{Guo2020ECCV}.
Specifically, for all compared attacks, we use SiamRPN++ (ResNet50) as the guidance to perform optimization or training. For example, we set $\phi_{\theta_t}$ in the objective function of OP-ABA (\ie, Eq.~\eqref{eq:adv_obj}) as the model of SiamRPN++ (ResNet50).
We report the precision drop after attacking in Table~\ref{tab:transfer} and the BRISQUE as the image quality assessment for generated adversarial frames.

As shown in Table \ref{tab:transfer}, we observe: \ding{182} Our methods, \ie, OP-ABA and OS-ABA, achieve the best and second-best transferability (\ie, higher precision drop) against DiMP50, DiMP18 \cite{Bhat2019ICCV}, and KYS \cite{bhat2020know}, hinting that our methods are more practical for black-box attacking. \ding{183} According to BRISQUE results, the adversarially blurred frames have smaller values than other adversarial examples, hinting that our methods are able to generate more natural frames since motion blur is a common degradation in the real world.



\begin{table}[t]
\centering
\scriptsize
\caption{Comparison results on transferability. Specifically, we use the adversarial examples crafted from SiamRPN++ (ResNet50) to attack four state-of-the-art trackers including SiamRPN++ (MobileNetv2) \cite{Li2019CVPR}, DiMP50 \cite{Bhat2019ICCV}, DiMP18 \cite{Bhat2019ICCV}, and KYS \cite{bhat2020know} on OTB100. We also calculate the average BRISQUE values of all adversarial examples.} \label{tab:transfer}
\begin{tabular}{l|c|c|c|c||c}
\toprule
    Trackers & \makecell{SiamRPN++\\(MobNetv2)} & DiMP50 & DiMP18 & KYS  & BRISQUE $\downarrow$ \\
\midrule
    Org. Prec.  & 86.4   & 89.2 & 87.1 & 89.5 &  20.15 \\
\midrule
    CSA & 0.2    & 3.4  &2.7 & 0.8 & 33.63\\
    SPARK  & \second{0.9}    & 2.0  & 1.0 & 0.9  & 24.78 \\
\midrule
    OP-ABA      &\first{2.5}   & \second{6.6}  & \second{10.3} & \second{7.9}  & \first{21.39}\\
    OS-ABA      & 0.2     &\first{10.7}  & \first{11.2} &\first{12.3} & \second{22.94} \\
\bottomrule
\end{tabular}\vspace{-5pt}
\end{table}




\section{Conclusion}\label{sec:concl}

In this work, we proposed a novel adversarial attack against visual object tracking, \ie, adversarial blur attack (ABA), considering the effects of motion blur instead of the noise against the state-of-the-art trackers. We first identified the motion blur synthesizing process during tracking based on which we proposed the optimization-based ABA (OP-ABA). This method fools the trackers by iteratively optimizing a tracking-aware objective but causes heavy time cost. We further proposed the one-step ABA by training a novel designed network to predict blur parameters in a one-step way. The attacking results on four public datasets, the visualization results, and comparison results demonstrated the effectiveness and advantages of our methods. This work not only reveals the potential threat of motion blur against trackers but also could work as a new way to evaluate the motion-blur robustness of trackers in the future.

\noindent\textbf{Acknowledgments}:
This work is supported in part by JSPS KAKENHI Grant No.JP20H04168, JP19K24348, JP19H04086, JP21H04877, JST-Mirai Program Grant No.JPMJMI20B8, Japan. Lei Ma is also supported by  Canada CIFAR AI Program and Natural Sciences and Engineering Research Council of Canada.
The work was also supported by the National Research Foundation, Singapore under its the AI Singapore Programme (AISG2-RP-2020-019), the National Research Foundation, Prime Ministers Office,
Singapore under its National Cybersecurity R\&D Program (No. NRF2018NCR-NCR005-0001), NRF Investigatorship NRFI06-2020-0001,
the National Research Foundation through its National Satellite of Excellence in Trustworthy Software
Systems (NSOE-TSS) project under the National Cybersecurity R\&D (NCR)
Grant (No.~NRF2018NCR-NSOE003-0001).
We gratefully acknowledge the support of NVIDIA AI Tech Center (NVAITC) to our research.

{\small
\bibliographystyle{ieee_fullname}
\bibliography{ref}
}


\begin{table*}[t]
\centering
\scriptsize
\caption{Attacking results of OP-ABA against KYS and DiMP with ResNet50 and ResNet18 as backbones on OTB100 and VOT2018, respectively.} \label{tab:supp_cmp_1}
\begin{tabular}{l|c|c|c|c|c|c|c}
\toprule
    \multirow{2}{*}{Backbones} & \multirow{2}{*}{Trackers}  &\multicolumn{4}{c|}{OTB100} & \multicolumn{2}{c}{VOT2018} \\
       &  &  Org.~Prec. & Prec.~Drop $\uparrow$ & Org.~Succ. & Succ.~Drop $\uparrow$ & Org. EAO  & EAO~Drop $\uparrow$ \\
\midrule
    \multirow{2}{*}{ResNet50} & KYS  & 89.5 & 18.8 & 68.6 & 13.8 & 0.405 & 0.289 \\
    & DiMP &89.2 & 30.9 & 68.9 & 23.6 & 0.423 & 0.405 \\
\midrule
     ResNet18 & DiMP  & 87.1 & 37.3 & 66.7 & 27.8 & 0.351 & 0.332 \\
\bottomrule
\end{tabular}
\end{table*}

\begin{table*}[t]
\centering
\scriptsize
\caption{Attacking results of OP-ABA against SiamRPN++ with ResNet50 and MobileNetv2 on UAV123 and LaSOT.
} \label{tab:supp_cmp_2}
\begin{tabular}{l|c|c|c|c|c|c|c|c|c}
\toprule
    \multirow{2}{*}{Backbones} & \multirow{2}{*}{Trackers}  & \multicolumn{4}{c|}{UAV123} & \multicolumn{4}{c}{LaSOT}   \\
       &  & Org. Prec. & Prec.~Drop. $\uparrow$ & Org.~Succ. & Succ.~Drop $\uparrow$ &  Org.~Prec. & Prec.~Drop $\uparrow$ & Org.~Succ. & Succ.~Drop $\uparrow$ \\
\midrule
    \multirow{2}{*}{ResNet50} & KYS  & 82.2 & 15.4 & 62.6 & 11.7 &  52.7 & 9.5 & 55.2 & 9.5 \\
    & DiMP& 84.4 & 32.4 & 63.9 & 24.6 & 54.4 & 21.1 & 55.3 & 18.7 \\
\midrule
    MobNetv2 & DiMP &81.0 & 39.0 & 61.5 & 29.9 &  51.5 & 25.2 &53.1 & 22.8\\
\bottomrule
\end{tabular}
\end{table*}

\newpage
\appendix

\section{Supplementary Material}

In this material, we report more results for attacking KYS \cite{bhat2020know} and DiMP \cite{Bhat2019ICCV}, correlation filtering (CF) and long-term trackers. We also provide extra ablation study by showing the loss values of OP-ABA w/o $\mathcal{A}$, OP-ABA w/o $\mathcal{W}$, and OP-ABA during optimization. In addition, we provide seven attacking cases with `gif' files in \href{https://tsingqguo.github.io/files/ICCV21MotAdvSupp.zip}{https://tsingqguo.github.io/files/ICCV21MotAdvSupp.zip}.

\subsection{Principle of Motion Blur}

A motion-blurred image is caused by the light accumulation during exposure and represented as \cite{Nah2017CVPR}
\begin{align} \label{eq:real_motionblur}
\mathbf{I}_t = \text{CRF}(\frac{1}{T}\int_{\tau=0}^{T}\mathbf{L}_t^{\tau})\simeq\text{CRF}(\frac{1}{N}\sum_{i=1}^{N}\mathbf{L}_t^{i}),
\end{align}
where $T$ is the exposure time, $\text{CRF}(\cdot)$ is the camera response function, and $\mathbf{I}_t^i=\text{CRF}(\mathbf{L}_t^i)$ where $\mathbf{L}_t^i$ is the raw light signal of $\mathbf{I}_t^i$. In practice, the CRF for spatial-variant blur is unknown \cite{Nah2017CVPR}. Hence, recent works simplify Eq.~(\ref{eq:real_motionblur}) as the averaging of neighboring instant frames \cite{Noroozi2017,Brooks2019CVPR,guo2021exloring}, \ie,
%
%
$\mathbf{I}_t = \frac{1}{N}\sum_{i=1}^{N}\mathbf{I}_t^{i}.$
%
Our synthesis method (Eq.~(8)) extends it to the spatial-variant accumulation via $\{\mathbf{A}_t^i\}$ and \textit{Hadamard product}, representing more challenging blur \cite{Nah2017CVPR,Guo2020NeurIPS}. \textit{Compared with the convolution} that reconstructs pixels via a weighted summation on their spatial neighbors, motion blur produces the blurred pixels by performing weighted summation across the instant images.

\subsection{Results of OP-ABA against Other Trackers}

In this material, we report more results of OP-ABA for attacking KYS and DiMP on the OTB100, UAV123, LaSOT, and VOT2018 datasets. As shown in Table~\ref{tab:supp_cmp_1} and \ref{tab:supp_cmp_2}, OP-ABA can reduce the precision and success rates of KYS, DiMP (ResNet50), and DiMP (ResNet18) on OTB100, VOT2018, UAV123, and LaSOT, significantly. When we compare the attack results of DiMP with those of KYS under the same backbone (\ie, ResNet50), it is easier for OP-ABA to attack DiMP since we achieve much higher precision or success rate drop. Compared DiMP (ResNet50) with DiMP (ResNet18), we see that the DiMP with deeper backbone is harder to be attacked since OP-ABA has lower performance drops on the DiMP (ResNet50), which is consist with the results reported in Table~1 and 2 in the main manuscript.

\subsection{Attacking CF and Long-term Trackers}
Although our method is tracker-aware, we can conduct transfer-based black-box attack by using the adversarial examples (AEs) from one tracker to attack another one.
As shown in Table~\ref{tab:transfer}, our method achieves the highest transferability, demonstrating the flexibility of our attacks.
We further conduct the transfer-based attack on traditional CF trackers, \eg, ECO and KCF. As shown in Table~\ref{tab:attack_cf},  OS-ABA is still effective for attacking CF trackers, showing higher transferability than the SOTA attacks, \ie, CSA and SPARK.
We further use OP-ABA to attack \textit{a long-term tracker (\ie, DiMP-LMUT \cite{Dai_2020_CVPR})}. As shown in Table~\ref{tab:long-term}, compared with the short-term tracker, the long-term tracker with global searching and meta-updater can defend against our attack to some extent and gets slightly smaller accuracy drop, hinting a potential defence method.
%
\begin{table}[h]
\footnotesize
\centering
\caption{Transfer-based black-box attack against a short-term tracker DiMP and its long-term version DiMP-LTMU on OTB100.}
\label{tab:long-term}
\resizebox{1.0\linewidth}{!}{
\begin{tabular}{l|cccc}
\toprule
Method & Org.~Succ. & Succ.~Drop $\uparrow$ & Org.~Prec. & Prec.~Drop $\uparrow$\\
\hline
DiMP      & 68.9 &  5.1 & 89.2 & 6.6 \\
DiMP-LMUT & 68.6 &  3.9 & 89.1 & 4.6 \\
\bottomrule
\end{tabular}
}
\end{table}
%

\begin{table}[h]
\centering
\caption{Attacking CF trackers via AEs from SiamRPN++ (ResNet50) on OTB100.}
\label{tab:attack_cf}
\resizebox{1.0\linewidth}{!}{
\begin{tabular}{l|cc|cccc}
\toprule
\multirow{2}{*}{Method} & \multirow{2}{*}{Feat.} & \multirow{2}{*}{Org.~Prec.} & \multicolumn{4}{c}{Prec.~Drop $\uparrow$} \\
& & & CSA & SPARK & OP-ABA & OS-ABA \\
\hline
ECO   & VGG & 89.6 & 1.6 & 0.9 & 1.9 & \first{2.3} \\
KCF   & HOG & 69.2 & 2.8 & 3.3 & 3.5 & \first{5.4} \\
\bottomrule
\end{tabular}
}
\end{table}
%

\subsection{Visualization of OP-ABA Optimization Process}\label{sec:loss}

\begin{figure}[t]
\centering
\includegraphics[width=1.0\linewidth]{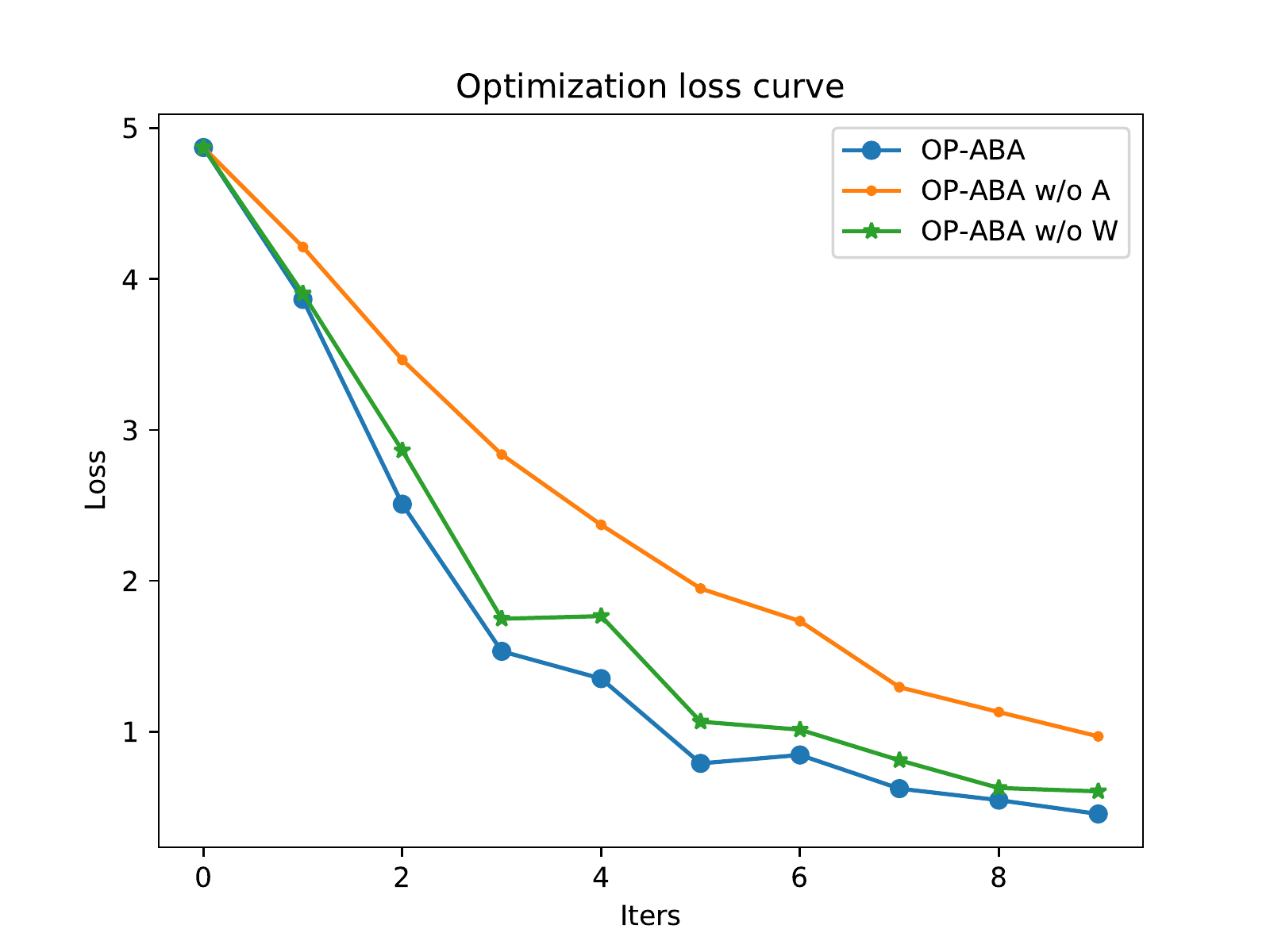}
   \caption{The optimization loss during the iteration of OP-ABA w/o $\mathcal{A}$, OP-ABA w/o $\mathcal{W}$, and OP-ABA.}
\label{fig:loss}\vspace{-0pt}
\end{figure}

In addition to the ablation study in the Sec~4.3 and Table~4, we further show the loss values of OP-ABA w/o $\mathcal{A}$, OP-ABA w/o $\mathcal{W}$, and OP-ABA during the iterative optimization in Fig.~\ref{fig:loss}. Clearly, the loss of OP-ABA considering both $\mathcal{A}$ and $\mathcal{W}$ reduces more quickly than other two variants. When we do not tune the $\mathcal{A}$ (\ie, OP-ABA w/o $\mathcal{A}$), the optimization process of OP-ABA w/o $\mathcal{A}$ becomes less effectively since the loss decreases slowly, demonstrating tunable $\mathcal{A}$ is significantly important for high attack success rate, which is consistent with the conclusion of Sec~4.3 and Table~4 in the main manuscript.

%
%

\subsection{Visualization of Attacking Cases}
We provide seven cases and compare the tracking results of SiamRPN++ (ResNet50) under the original frames, normal blurred frames, and adversarially blurred frames, respectively. Please find the `gif' files in \href{https://tsingqguo.github.io/files/ICCV21MotAdvSupp.zip}{https://tsingqguo.github.io/files/ICCV21MotAdvSupp.zip}. Clearly, the adversarially blurred frames have similar appearance with the normal one. The tracker is robust to normal blur but is easily affected by the proposed adversarially blurred frames.


\end{document}